\documentclass[11pt,a4paper]{article}

\usepackage[margin=1in]{geometry}
\usepackage{hyperref}
\usepackage{booktabs}
\usepackage{natbib}
\usepackage{graphicx}
\usepackage{amsmath}
\usepackage{amssymb}
\usepackage{url}
\usepackage[protrusion=true,expansion=false]{microtype}

\hypersetup{colorlinks=true,linkcolor=black,urlcolor=blue,citecolor=blue}

\newcommand{\plusminus}{$\pm$}

\title{ConvMemory: A Lightweight Learned Memory Reranker, \\
       a Negative Attribution Result, \\
       and a Research-Preview Conflict Editor}

\author{
Taiheng Pan \\
School of Computing and Information Systems \\
University of Melbourne \\
\texttt{github.com/pth2002}
}

\date{\today}

\begin{document}

\maketitle

\begin{abstract}
We describe ConvMemory, a small ($\sim$3.6M-parameter) learned reranker for conversational long-term memory retrieval, trained with cross-encoder teacher supervision over fused dense and lexical features. On the LongMemEval memory family, ConvMemory operates above the BGE-large cross-encoder in Recall@10 at 12--47$\times$ lower latency, remains within 0.025 Recall@10 of mxbai-rerank-large-v1 on Clean500 while running 28$\times$ cheaper; under Stress1000 distractors the Recall@10 gap widens to 0.081 but ConvMemory still operates at 117$\times$ lower latency; these LongMemEval numbers are single-run or single-seed and are reported as indicative cost-frontier evidence, not benchmark-grade. We then publish a rigorous negative attribution result on a previously claimed mechanism: a five-seed retrained ablation with paired bootstrap shows that ConvMemory's learned temporal window is statistically significant on aggregate but \emph{not} temporally specific, with the largest effects on hard non-temporal controls and no significant effect on multi-hop temporal queries. The honest description of the mechanism is cheap cross-encoder distillation in a fused dense+lexical feature space, not temporal-structure exploitation. We additionally release CCGE-LA, a low-amplitude conflict-aware candidate-set editor over ConvMemory, as a research preview with modest but consistent gains on supersession and stale/rescue slices on LoCoMo. All results are retrieval-stage; ConvMemory does not match mxbai-rerank-large-v1 in absolute LoCoMo MRR, and the report is single-author and not yet independently audited.
\end{abstract}

\section{Introduction}
\label{sec:intro}

Long-term agent and conversational memory systems accumulate large memory stores over time. Per-turn LLM-as-selector approaches scale poorly: their cost grows with the number of memories $N$, and aggressive truncation strategies discard recall. Strong cross-encoder rerankers are accurate but expensive at memory-store scale, often spending hundreds of milliseconds to several seconds per query when run over the top-500 candidates of a dense retriever.

This report studies one narrow question. How cheaply can a small learned reranker approximate cross-encoder quality on memory-style retrieval, and what mechanism is responsible for any quality it does deliver?

We make three contributions of varying maturity. We list them as contributions rather than promote them as equal main results: the first two are audited multi-seed results, while the third is an early-stage extension distributed as an alpha checkpoint.

\begin{enumerate}
    \item \textbf{ConvMemory} (engineering contribution): a small ($\sim$3.6M-parameter) learned reranker with cross-encoder teacher supervision, distributed via the Hugging Face Hub. Its practical value is a cost-quality operating point: roughly cross-encoder Recall@10 on conversational memory retrieval at one to two orders of magnitude lower latency. We do not claim it beats strong cross-encoders generally; in particular, mxbai-rerank-large-v1 dominates it on absolute LoCoMo MRR.
    \item \textbf{A rigorous attribution study} (scientific contribution): showing that the originally proposed temporal-window mechanism is not load-bearing on this task. The gain is best described as cheap cross-encoder distillation in a fused dense+lexical feature space. We publish this as a negative result on a previously claimed mechanism because the engineering value (cost-effective reranking) survives the negative attribution.
    \item \textbf{CCGE-LA} (research preview): a low-amplitude conflict-aware candidate-set editor over ConvMemory. It shows modest but consistent gains on stale/current conflict slices on LoCoMo. The distributed checkpoint is single-seed; the aggregate gains rest on an internal 5-seed study. We list this as a contribution to make the published artifact and its limits explicit, not to claim equal maturity with the other two.
\end{enumerate}

Throughout, we use the phrase ``to our knowledge'' rather than ``first'' and avoid blanket claims of novelty or state-of-the-art performance. Each section that reports a number also states what that number is \emph{not} evidence for.

\section{Related Work}

\paragraph{Dense memory retrieval.} ConvMemory is built on top of standard sentence encoders. We use MPNet \citep{mpnet2020} as the default backbone for the distributed checkpoint and additionally retrain on BGE-large \citep{bge_reranker2023} and E5-large \citep{e5_2022} encoders to test backbone generalization.

\paragraph{Cross-encoder rerankers.} The strong reranker baselines we compare against are ms-marco-MiniLM-L-6 \citep{msmarco_minilm}, the BGE-reranker family \citep{bge_reranker2023}, jina-reranker-v2-multilingual \citep{jina_reranker2024}, and mxbai-rerank-large-v1 \citep{mxbai_rerank2024}. Cross-encoder distillation into smaller students is a well-known recipe; ConvMemory inherits that approach with a fused dense+lexical feature space rather than raw text--text concatenation.

\paragraph{Agent memory systems.} Recent agent memory frameworks such as mem0 \citep{mem0_2024} and Letta/MemGPT \citep{letta2023} motivate the practical setting: a long-running agent must retrieve relevant fragments from an accumulated memory store at every turn, often under tight latency budgets.

\paragraph{Benchmarks.} We use LoCoMo \citep{locomo2024} as the primary conversational memory benchmark and LongMemEval \citep{longmemeval2024} for cost-frontier measurements. External out-of-distribution scope checks use QMSum \citep{qmsum2021}, MSC persona \citep{msc2022}, HotpotQA \citep{hotpotqa2018}, and MuSiQue \citep{musique2022}.

\section{ConvMemory Architecture}
\label{sec:architecture}

\subsection{Configuration}

Table~\ref{tab:config} summarizes the distributed ConvMemory checkpoint. The total parameter count is approximately 3.6M.

\begin{table}[ht]
\centering
\caption{ConvMemory checkpoint configuration (the distributed LoCoMo MPNet checkpoint).}
\label{tab:config}
\begin{tabular}{lr}
\toprule
Field & Value \\
\midrule
Embedding backbone & sentence-transformers/all-mpnet-base-v2 \\
Embedding dimension & 768 \\
Window size & 5 \\
Stride & 1 \\
Kernel size & 3 \\
Hidden dimension & 256 \\
Token MLP dim & 32 \\
Channel MLP dim & 512 \\
Candidate top-$n$ & 500 \\
Raw score fusion weight & 0.025 \\
Total parameters & $\sim$3{,}648{,}118 \\
\bottomrule
\end{tabular}
\end{table}

\subsection{Components}

ConvMemory composes four components into a single CE-lite scoring function for each candidate in the dense retriever's top-500 pool.

\begin{enumerate}
    \item \textbf{Dense candidate scoring.} The base sentence encoder scores query-candidate cosine similarity over the memory store; the top-500 candidates by dense score form the pool.
    \item \textbf{Sliding-window conv/mixer encoder.} A small windowed encoder (window size 5, stride 1, kernel size 3, hidden dim 256, token-MLP 32, channel-MLP 512) processes the candidate stream within the top-500 pool. \emph{The window operates over the candidate ranking order (sorted by dense score), not over chronological memory order.} Early versions of this work conjectured that, because candidate ranking and memory time are correlated in conversational data, the windowed encoder would indirectly capture temporal structure. Section~\ref{sec:attribution} shows that this interpretation does not survive a proper attribution study.
    \item \textbf{Lexical interaction features.} Query-candidate lexical features (token overlap and related) provide a cheap, discriminative signal that complements the dense channel.
    \item \textbf{Optional block-level router scalar.} A learned scalar gates between sub-features. Its contribution is approximately zero in the retrained ablation (Section~\ref{sec:attribution}).
\end{enumerate}

\subsection{Training objective}

ConvMemory is trained with cross-encoder teacher supervision on LoCoMo top-500 candidate pools. The teacher is \texttt{cross-encoder/ms-marco-MiniLM-L-6-v2}. We deliberately do not match the teacher to the strongest available cross-encoder (mxbai-rerank-large-v1); doing so would conflate distillation gains with teacher choice and weaken the cost-frontier comparison. The objective combines a pairwise ranking loss over candidate pairs with a first-rank objective that pushes the gold memory to rank 1.

\subsection{Cost structure}

The structural source of ConvMemory's cost advantage is feature pre\-computability on the candidate side. Dense embeddings and candidate-side lexical statistics (e.g., candidate tokenization, IDF-weighted term sets) are computed at memory-write time and cached. The truly query-dependent lexical interaction features (e.g., token overlap, query-conditioned weighting) are computed at inference time, but cheaply: they are sparse set operations over the cached candidate-side tokenization, not transformer forward passes. The remaining inference work is a small per-query fusion --- the conv/mixer encoder over the 500-candidate window and the CE-lite head. The cross-encoder baselines we compare against, in contrast, must run a full transformer forward over each query-candidate pair at query time, which cannot be amortized across queries.

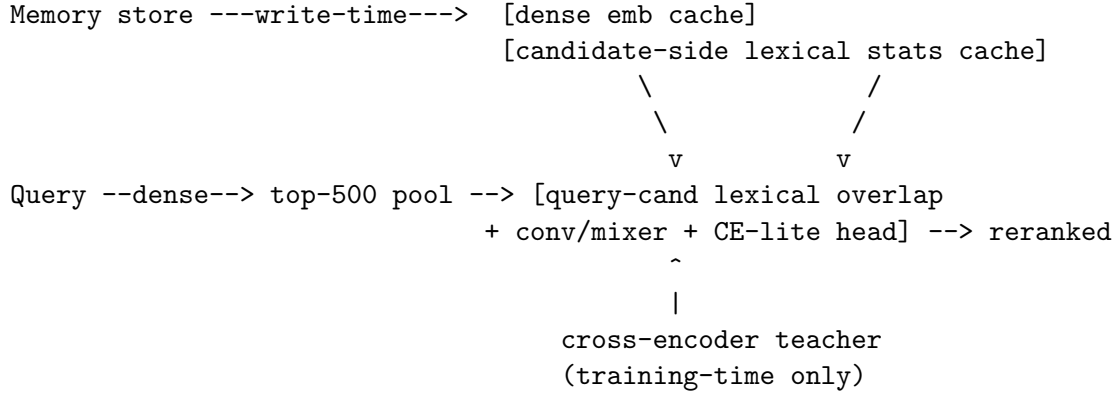
\begin{verbatim}
   Memory store ---write-time--->  [dense emb cache]
                                   [candidate-side lexical stats cache]
                                            \              /
                                             \            /
                                              v          v
   Query --dense--> top-500 pool --> [query-cand lexical overlap
                                  + conv/mixer + CE-lite head] --> reranked
                                              ^
                                              |
                                       cross-encoder teacher
                                       (training-time only)
\end{verbatim}

We do not claim this architecture is novel; the contribution is the cost-quality operating point this particular combination delivers on conversational memory retrieval, and the careful attribution study of why it works.

\section{Main Results}

\subsection{LongMemEval cost advantage}
\label{sec:longmemeval}

Table~\ref{tab:longmemeval} reports Recall@10, MRR, and per-query latency on LongMemEval. Two settings are reported: Clean500 (no distractor stress) and Stress1000 with seed 23 (1000 distractor memories per question).

\begin{table}[ht]
\centering
\caption{LongMemEval cost-frontier measurements. Clean500 is single-run; Stress1000 is single-seed (seed 23). Indicative cost-frontier evidence, not benchmark-grade.}
\label{tab:longmemeval}
\begin{tabular}{llrrr}
\toprule
Setting & Method & Recall@10 & MRR & ms/query \\
\midrule
Clean500, BGE-large CE   & Raw MPNet              & 0.9049 & 0.7829 &    0.01 \\
Clean500, BGE-large CE   & BGE-large CE top500    & 0.8807 & 0.8574 &  555.69 \\
Clean500, BGE-large CE   & ConvMemory top500      & 0.9593 & 0.8973 &   44.00 \\
Clean500, mxbai CE       & mxbai CE top500        & 0.9835 & 0.9317 & 1129.14 \\
Clean500, mxbai CE       & ConvMemory top500      & 0.9593 & 0.8973 &   40.80 \\
Stress1000 s23, BGE CE   & BGE-large CE top500    & 0.6913 & 0.6651 & 5231.77 \\
Stress1000 s23, BGE CE   & ConvMemory cand-local  & 0.7386 & 0.6125 &  110.71 \\
Stress1000 s23, mxbai CE & mxbai CE top500        & 0.8195 & 0.7044 & 11211.63 \\
Stress1000 s23, mxbai CE & ConvMemory cand-local  & 0.7386 & 0.6125 &   95.57 \\
\bottomrule
\end{tabular}
\end{table}

On Clean500, ConvMemory has higher Recall@10 than the BGE-large cross-encoder (0.9593 vs 0.8807) at $555.69 / 44.00 \approx 12.6\times$ lower latency, and stays within 0.025 Recall@10 of mxbai-rerank-large-v1 (0.9593 vs 0.9835) at $1129.14 / 40.80 \approx 27.7\times$ lower latency. Under Stress1000 distractors, ConvMemory beats the BGE-large cross-encoder on Recall@10 (0.7386 vs 0.6913) at $5231.77 / 110.71 \approx 47.3\times$ lower latency. Against mxbai-rerank-large-v1 under Stress1000, ConvMemory's Recall@10 drops 0.081 below mxbai (0.7386 vs 0.8195), but it runs $11211.63 / 95.57 \approx 117.3\times$ cheaper; we report this trade-off explicitly rather than presenting only the favorable Clean500 comparison.

\paragraph{Latency measurement protocol (summary).} All ms/query figures are reranking-only wall-clock time over the top-500 candidate pool, measured on a single NVIDIA GeForce RTX 4080 SUPER (32GB). Cross-encoder batch sizes are tuned per model to a throughput-favourable setting (32--128 query-candidate pairs per forward, larger for smaller models); ConvMemory's per-query fusion runs at query batch size 1 with GPU-vectorized scoring of the 500-candidate window. Dense embedding computation, candidate-side feature precomputation, and dense retrieval are excluded from all rows --- they are shared across methods and would only shift all numbers by the same constant. The Raw MPNet 0.01 ms/query figure reflects only the cost of sorting precomputed dense scores; it is included as a lower bound, not as a comparable method. Full protocol, including hardware specification, software stack, per-model batching configuration, and what is and is not included in each measurement, is in Appendix~\ref{app:latency}.

\paragraph{What this is not evidence for.} These LongMemEval numbers are single-run (Clean500) or single-seed (Stress1000 seed 23). They are intended as a cost-frontier indicator on memory-style retrieval, not as a benchmark-grade comparison. On MRR specifically, ConvMemory loses to both cross-encoders on Stress1000.

\subsection{LoCoMo strong cross-encoder baselines}
\label{sec:locomo_baselines}

Table~\ref{tab:locomo} reports a 5-seed comparison against strong cross-encoder rerankers on LoCoMo, with seeds 7, 11, 23, 31, 47 and the dense retriever's top-500 as input pool.

\begin{table}[ht]
\centering
\caption{LoCoMo strong reranker baselines (5 seeds: 7, 11, 23, 31, 47; raw dense top-500 pool). All entries are 5-seed mean \plusminus{} std. The ConvMemory v0.40 row aggregates per-seed metrics from the archived v0.40 evaluation; earlier public README summaries conservatively omitted the Hit@10 and MRR variance, but the per-seed CSV retains the full distribution and is the source for this row.}
\label{tab:locomo}
\begin{tabular}{lrrr}
\toprule
Reranker & Recall@10 & Hit@10 & MRR \\
\midrule
ConvMemory (v0.40, 5 seeds)         & 0.7798 \plusminus 0.0074 & 0.8350 \plusminus 0.0083 & 0.5824 \plusminus 0.0189 \\
BGE-reranker-base                   & 0.6967 \plusminus 0.0126 & 0.7469 \plusminus 0.0144 & 0.5469 \plusminus 0.0140 \\
Jina-reranker-v2-base-multilingual  & 0.7411 \plusminus 0.0103 & 0.7924 \plusminus 0.0083 & 0.5754 \plusminus 0.0074 \\
BGE-reranker-large                  & 0.7621 \plusminus 0.0155 & 0.8124 \plusminus 0.0135 & 0.6120 \plusminus 0.0144 \\
mxbai-rerank-large-v1               & 0.8080 \plusminus 0.0153 & 0.8486 \plusminus 0.0108 & 0.6687 \plusminus 0.0093 \\
\bottomrule
\end{tabular}
\end{table}

\paragraph{Honest reading.} ConvMemory is competitive with the BGE-reranker family on Recall@10 (above BGE-reranker-base and BGE-reranker-large in mean Recall@10 while being one to two orders of magnitude smaller in parameter count depending on the baseline) but \emph{loses} to mxbai-rerank-large-v1 on both Recall@10 and MRR. We do not aggregate ConvMemory and CCGE-LA gains together to argue against mxbai: even with the CCGE-LA editor on top, the absolute MRR remains well below mxbai-rerank-large-v1.

\subsection{Strong-backbone retraining}

Table~\ref{tab:backbones} reports a 3-seed retraining study with stronger embedding backbones (BGE-large and E5-large) in place of MPNet.

\begin{table}[ht]
\centering
\caption{Strong-backbone retraining on LoCoMo (3 seeds). The +0.09--0.10 Recall@10 lift survives when the backbone is swapped.}
\label{tab:backbones}
\begin{tabular}{lrrrr}
\toprule
Backbone & Raw R@10 & ConvMemory R@10 & Gain & ConvMemory MRR \\
\midrule
BGE-large & 0.6680 \plusminus 0.0237 & 0.7726 \plusminus 0.0100 & +0.1046 \plusminus 0.0137 & 0.5639 \plusminus 0.0066 \\
E5-large  & 0.7010 \plusminus 0.0216 & 0.7902 \plusminus 0.0171 & +0.0892 \plusminus 0.0052 & 0.5941 \plusminus 0.0103 \\
\bottomrule
\end{tabular}
\end{table}

\paragraph{Reading.} Gains generalize across embedding backbones. \textbf{Caveat:} 3-seed study, and the BGE-large / E5-large ConvMemory checkpoints are reported here but not currently distributed via the Hugging Face Hub.

\section{Rigorous Attribution: A Negative Result}
\label{sec:attribution}

This section is the headline scientific contribution of the report. We begin by stating the original conjecture, then present the evidence that refutes it.

\paragraph{Original conjecture.} As stated in Section~\ref{sec:architecture}, the windowed conv/mixer encoder operates over candidate \emph{ranking} order, not chronological order. Early versions of this work conjectured that, because dense-rank neighborhoods on conversational memory data are correlated with temporal locality (recent and topically-similar memories tend to cluster together in the dense ranking), the windowed encoder would \emph{indirectly} learn temporal structure useful for memory retrieval, and that this implicit temporal capture was the load-bearing reason ConvMemory worked on T\_SUP and T\_HOP query slices. The evidence below shows this interpretation does not survive a careful study: the window helps, but not on the slices a temporal mechanism would help on.

\subsection{Retrained ablation (3 seeds)}

Table~\ref{tab:ablation3} reports a 3-seed retrained ablation on MPNet. Each variant is retrained from scratch with the corresponding component removed.

\begin{table}[ht]
\centering
\caption{Retrained ablation, 3 seeds, MPNet backbone. Lexical features dominate; the temporal window is smaller; the router is approximately zero.}
\label{tab:ablation3}
\begin{tabular}{lrrr}
\toprule
Variant & Recall@10 & MRR & $\Delta$ R@10 vs full \\
\midrule
full\_control          & 0.7474 \plusminus 0.0229 & 0.5343 \plusminus 0.0160 & 0.0000                   \\
no\_router             & 0.7491 \plusminus 0.0213 & 0.5391 \plusminus 0.0137 & +0.0017 \plusminus 0.0020 \\
no\_temporal\_w1       & 0.7121 \plusminus 0.0232 & 0.5305 \plusminus 0.0148 & $-$0.0353 \plusminus 0.0052 \\
no\_lexical            & 0.6584 \plusminus 0.0185 & 0.4367 \plusminus 0.0129 & $-$0.0890 \plusminus 0.0061 \\
no\_lexical\_no\_router & 0.6574 \plusminus 0.0163 & 0.4342 \plusminus 0.0127 & $-$0.0899 \plusminus 0.0087 \\
\bottomrule
\end{tabular}
\end{table}

The lexical features contribute the largest delta ($-$0.089 Recall@10 when removed). The temporal window contributes a smaller but non-trivial delta ($-$0.035). The router scalar contributes approximately zero (+0.0017 \plusminus 0.0020). This alone is insufficient to claim what the temporal contribution \emph{is}, only that it is not negligible on aggregate.

\subsection{Five-seed retrained attribution with paired bootstrap}

We then ran the v0.51 5-seed retrained attribution and computed paired bootstrap confidence intervals for the full\_control $-$ no\_temporal\_w1 delta on Recall@10, sliced by query type.

Mean Recall@10 (5 seeds): full\_control\_retrained 0.7432 \plusminus 0.0207, no\_temporal\_w1\_retrained 0.7054 \plusminus 0.0221, tuned\_heuristic (dense+BM25+temporal+decay) 0.7234 \plusminus 0.0227, raw\_dense 0.5345 \plusminus 0.0210.

\begin{table}[ht]
\centering
\caption{Paired bootstrap deltas, full\_control $-$ no\_temporal\_w1, Recall@10. The aggregate delta is significant on all slices except T\_HOP\_auto; the rank ordering across slices is the reverse of what a temporal-specific mechanism would predict (see Section~\ref{sec:attribution} Interpretation).}
\label{tab:bootstrap}
\begin{tabular}{lrrl}
\toprule
Slice & $\Delta$ & 95\% CI & Reading \\
\midrule
ALL                       & +0.0376 & [+0.0306, +0.0451] & significant \\
T\_SUP\_auto              & +0.0407 & [+0.0219, +0.0603] & significant \\
T\_REQUIRED\_auto         & +0.0252 & [+0.0139, +0.0363] & significant \\
T\_HOP\_auto              & +0.0096 & [$-$0.0037, +0.0230] & NOT significant \\
OTHER                     & +0.0868 & [+0.0672, +0.1045] & significant (largest) \\
HARD\_NON\_TEMPORAL\_auto & +0.0838 & [+0.0650, +0.1040] & significant \\
\bottomrule
\end{tabular}
\end{table}

Additionally, full\_control $-$ tuned\_heuristic on ALL Recall@10 is +0.0199 [+0.0105, +0.0283], and the corresponding MRR delta is +0.0566.

\paragraph{Interpretation.} The aggregate gain attributed to the window is statistically significant, and on individual slices the window does help on T\_SUP\_auto (+0.0407) and T\_REQUIRED\_auto (+0.0252); these are significant and are by themselves compatible with some temporal-relevant signal. The slice pattern as a whole, however, does not match what a temporal-specific mechanism would predict. A genuinely temporal mechanism should produce its largest effects on the most temporally-demanding slice (multi-hop temporal reasoning, \texttt{T\_HOP\_auto}) and smaller or negligible effects on hard non-temporal controls. The observed pattern is the reverse rank ordering: the effect is \emph{largest} on \texttt{HARD\_NON\_TEMPORAL\_auto} (+0.0838) and \texttt{OTHER} (+0.0868), and is \emph{not} statistically significant on \texttt{T\_HOP\_auto} (+0.0096, CI crosses zero).

The fact that the window helps \emph{more} on hard non-temporal queries than on temporal-hop queries means the mechanism is not uniquely or specifically temporal. The most parsimonious description is that the windowed encoder operates as generic neighborhood / capacity smoothing in the fused dense+lexical feature space, with T\_SUP remaining an open slice where some temporal-relevant signal may still be present but cannot be cleanly attributed to the window from this experiment alone. The overall mechanism best fits cheap cross-encoder distillation in the fused dense+lexical feature space, with the windowed encoder providing modest extra capacity that helps broadly on hard queries.

\paragraph{Why publish this.} We publish this negative result on a previously claimed mechanism because the discipline matters. The engineering value of ConvMemory (cost-effective reranking at a useful operating point) is independent of the mechanism story and survives the negative attribution.

\paragraph{What this is not evidence for.} This negative result is about \emph{ConvMemory's} temporal window on \emph{LoCoMo}. It does not rule out the possibility that other temporal mechanisms in other architectures on other benchmarks are load-bearing. It also does not rule out the possibility that a different temporal mechanism, properly designed, would help on T\_HOP.

\section{CCGE-LA: Research-Preview Conflict Editor}

CCGE-LA (``Low-Amplitude Counterfactual Conflict Graph Editor'') is a tiny editor over ConvMemory's candidate set, released as a research preview.

\paragraph{Motivation.} As shown later in Section~\ref{sec:v059}, small from-scratch stream rerankers on real LoCoMo fail without a teacher signal. CCGE-LA takes the opposite approach: rather than replacing ConvMemory with a new from-scratch model, it reads ConvMemory's candidate set and applies a small residual correction when the candidate set looks conflict-prone (e.g., when a stale and a current memory both score highly).

\subsection{Architecture}

CCGE-LA computes 18 candidate-set features for each candidate in the top-$k$ pool:

\begin{quote}
\sloppy
\texttt{base\_score\_z},
\texttt{dense\_score\_z},
\texttt{position\_z},
\texttt{query\_overlap\_z},
\texttt{base\_rank\_norm},
\texttt{dense\_rank\_norm},
\texttt{sim\_to\_base\_top},
\texttt{sim\_to\_dense\_top},
\texttt{semantic\_density\_top16},
\texttt{token\_overlap\_to\_top},
\texttt{newer\_than\_base\_top},
\texttt{older\_than\_base\_top},
\texttt{abs\_pos\_gap\_top\_z},
\texttt{base\_margin\_1\_2},
\texttt{base\_entropy\_top16},
\texttt{conflict\_density\_top16},
\texttt{time\_span\_top16},
\texttt{top\_overlap}.
\end{quote}

These features are passed through a small transformer encoder, then a residual head produces a per-candidate score correction. A sigmoid gate with negative-bias initialization controls editor amplitude, so an untrained editor is approximately a no-op (gate close to 0). This is a deliberate safety property: the editor cannot make things much worse before it has learned anything.

\subsection{Training}

Training uses retrieval cross-entropy only. We explicitly enforce three anti-shortcut rules:

\begin{itemize}
    \item No current/stale labels in features. The editor does not get to peek at which candidate is canonical.
    \item No gold-defining feature. No feature uniquely identifies the gold answer.
    \item No distillation as the defining mechanism. The editor is not trained against a stronger teacher; it must learn from retrieval supervision alone.
\end{itemize}

\subsection{Results}

\begin{table}[ht]
\centering
\caption{CCGE-LA per-slice best from internal 5-seed exploration on real LoCoMo. Each row reports the strongest internal editor variant on that specific slice; the variants differ across rows (FULL is best under V150, T\_SUP\_auto under V145, the rescue slices under V149). This is an internal upper-bound frontier across our exploration, not the behavior of a single deployable model. The released alpha checkpoint, which is a single variant and a single seed, is reported separately in Table~\ref{tab:ccgela_alpha} and underperforms this per-slice frontier by design.}
\label{tab:ccgela}
\resizebox{\textwidth}{!}{%
\begin{tabular}{lrrr}
\toprule
Setting                          & ConvMemory MRR & Best editor MRR & $\Delta$ \\
\midrule
FULL                             & 0.5708 & 0.5878 & +0.0170 \\
T\_SUP\_auto                     & 0.5518 & 0.5803 & +0.0285 \\
CONV\_TOP1\_WRONG\_GOLD\_IN\_POOL & 0.2586 & 0.3285 & +0.0699 \\
RESCUABLE\_STALE\_TOP1           & 0.2636 & 0.3655 & +0.1019 \\
\bottomrule
\end{tabular}%
}
\end{table}

The lifts in Table~\ref{tab:ccgela} are not the behavior of a single CCGE-LA model. We ran multiple editor variants during internal exploration; different slices were strongest under different variants. Table~\ref{tab:ccgela} reports the per-slice best across this exploration. Read this way, the per-slice frontier shows: on FULL MRR the best variant (V150) lifts +0.0170 over ConvMemory base; on the supersession slice (T\_SUP\_auto) the best variant (V145) lifts +0.0285; on diagnostic conflict slices the best variants (V149) lift +0.0699 and +0.1019. The released alpha checkpoint (Table~\ref{tab:ccgela_alpha}) is a single variant and seed and lifts a smaller, uniform amount across slices. We present the per-slice frontier as evidence that conflict-aware editing carries useful signal under our slice diagnostics; we do not claim a single deployable model matches the frontier.

\begin{table}[ht]
\centering
\caption{CCGE-LA published alpha checkpoint (single seed, seed 23). The distributed checkpoint underperforms the best 5-seed editor variant; the 5-seed numbers in Table~\ref{tab:ccgela} should be read as aggregate gains, not as the released checkpoint's behavior.}
\label{tab:ccgela_alpha}
\begin{tabular}{lrrr}
\toprule
Subset                            & MRR & R@10 & gate \\
\midrule
FULL                             & 0.5638 & 0.7725 & 0.0995 \\
T\_SUP\_auto                     & 0.5508 & 0.7138 & 0.0995 \\
CONV\_TOP1\_WRONG\_GOLD\_IN\_POOL & 0.2994 & 0.6822 & 0.0995 \\
RESCUABLE\_STALE\_TOP1           & 0.3093 & 0.6877 & 0.0995 \\
\bottomrule
\end{tabular}
\end{table}

\paragraph{Honest framing.} Modest aggregate lift; larger lift on supersession and rescue slices. The published alpha checkpoint is single-seed. \textbf{This is explicitly a research preview / alpha.} The limits are: single task family (LoCoMo conversational memory), single backbone (MPNet), no broad cross-task validation. CCGE-LA is listed in this report as a research-preview contribution (see Section~\ref{sec:intro}); we do not claim it as a general conflict-resolution mechanism or as equal in maturity to the engineering and attribution contributions.

\section{Real-Data Stream-Reranker Negative Result (v0.59)}
\label{sec:v059}

A natural counter-question to ConvMemory is: do you really need a cross-encoder teacher, or can a small stream-architecture reranker learn this directly from retrieval supervision? We tested both a recurrent running-state reranker and a flat transformer over the memory stream, both trained from scratch on real LoCoMo with retrieval-only supervision. To our knowledge, this specific negative result has not been published before; it is reported here for the first time.

\begin{table}[ht]
\centering
\caption{v0.59 real-LoCoMo stream-reranker study, 5-seed full run. Main finding: both from-scratch stream rerankers fail below raw dense on the supersession slice; only the cross-encoder-distilled ConvMemory checkpoint beats raw dense. Cost ratio = inference time at large $N$ / inference time at small $N$; the recurrent state is structurally roughly flat in $N$ by construction. Full latency protocol in Appendix~\ref{app:latency}.}
\label{tab:v059}
\resizebox{\textwidth}{!}{%
\begin{tabular}{lrrrrrr}
\toprule
Arm & T\_SUP R@10 & T\_SUP MRR & FULL R@10 & FULL MRR & params & cost ratio \\
\midrule
raw\_dense                & 0.5005 \plusminus 0.0496 & 0.2939 \plusminus 0.0267 & 0.5345 \plusminus 0.0210 & 0.3254 \plusminus 0.0105 & 0          & 484.50$\times$ \\
recency\_rule             & 0.5275 \plusminus 0.0398 & 0.4109 \plusminus 0.0477 & 0.5318 \plusminus 0.0083 & 0.4108 \plusminus 0.0138 & 0          & 49.78$\times$  \\
strong\_reranker\_ref     & 0.7061 \plusminus 0.0306 & 0.5396 \plusminus 0.0293 & 0.7498 \plusminus 0.0080 & 0.5610 \plusminus 0.0116 & 3{,}648{,}118 & 7.99$\times$   \\
transformer\_over\_stream & 0.3127 \plusminus 0.0548 & 0.1681 \plusminus 0.0191 & 0.3638 \plusminus 0.0534 & 0.1927 \plusminus 0.0245 & 165{,}089   & 10.21$\times$  \\
running\_state\_gru       & 0.2826 \plusminus 0.0483 & 0.1379 \plusminus 0.0236 & 0.3474 \plusminus 0.0339 & 0.1787 \plusminus 0.0198 & 161{,}562   & 1.60$\times$   \\
running\_state\_gru\_large & 0.2953 \plusminus 0.0493 & 0.1503 \plusminus 0.0300 & 0.3414 \plusminus 0.0185 & 0.1831 \plusminus 0.0092 & 928{,}929   & 1.07$\times$   \\
\bottomrule
\end{tabular}%
}
\end{table}

The \texttt{strong\_reranker\_ref} arm is the ConvMemory checkpoint. ``Cost ratio'' is the inference cost at large $N$ relative to small $N$ (lower is better; recurrent state is structurally roughly flat in $N$).

\paragraph{Reading.} Both from-scratch stream rerankers fail below raw dense on the supersession slice (T\_SUP R@10 0.31 and 0.28 vs raw dense 0.50). Increasing capacity from 161K to 929K parameters does not rescue them. The recurrent state's cost asymmetry (roughly flat in $N$) is structurally real, but combined with the effectiveness failure it supports the conclusion that the teacher signal source is necessary at this data and parameter scale. ConvMemory's value is not an arbitrary stream-architecture effect.

\paragraph{What this is not evidence for.} It does not rule out larger-data or larger-parameter regimes where a from-scratch stream reranker might succeed. It rules out the small-budget, retrieval-only-supervised, real-LoCoMo regime studied here.

\section{Synthetic Benchmark Trajectory: A Methodology Lesson}

Across multiple internal synthetic benchmarks (versions roughly v0.52 through v1.42), benchmark design was repeatedly degenerate. In each iteration, the gold answer turned out to be recoverable by a trivial baseline: recency, position-equals-time, or oracle-equivalent supervised detectors. The cleanest current benchmark is V142, which was externally verifier-audited per \texttt{docs/RESEARCH\_TRAJECTORY.md} in the public repository.

Even on V142, the benchmark decomposes into two subsets that must be reported separately.

\begin{table}[ht]
\centering
\caption{V142 synthetic, current subset (no date in the query). On this subset, bitemporal is mathematically identical to event\_time\_no\_asof, because the as-of filter is skipped when the query has no date. The real mechanism on current queries is ``learned support extraction + event-time ordering.''}
\label{tab:v142current}
\begin{tabular}{lrrr}
\toprule
Arm & R@1 & MRR & stale@1 \\
\midrule
support\_only         & 0.4660 & 0.6909 & 0.3990 \\
event\_time\_no\_asof & 0.7360 & 0.8539 & 0.0700 \\
write\_order          & 0.7090 & 0.8123 & 0.1290 \\
bitemporal            & 0.7360 & 0.8539 & 0.0700 \\
\bottomrule
\end{tabular}
\end{table}

\begin{table}[ht]
\centering
\caption{V142 synthetic, as-of subset (date in the query; the query embeds the target date). The as-of gain depends on a hand-designed bitemporal ranking rule that uses the query-embedded date. This is not a learned mechanism on this subset.}
\label{tab:v142asof}
\begin{tabular}{lrrr}
\toprule
Arm & R@1 & MRR & stale@1 \\
\midrule
support\_only         & 0.6675 & 0.8025 & 0.3325 \\
event\_time\_no\_asof & 0.1450 & 0.5625 & 0.8550 \\
write\_order          & 0.3025 & 0.5771 & 0.6975 \\
bitemporal            & 0.9475 & 0.9738 & 0.0525 \\
\bottomrule
\end{tabular}
\end{table}

The honest claim is two separate results, not one aggregate ``bitemporal +0.27'' gain. The bitemporal arm consists of a learned operation/support extractor plus a hand-designed bitemporal ranking rule; we deliberately do not call it ``a learned bitemporal mechanism.''

\paragraph{Methodology lesson.} Cleanly attributing a mechanism on synthetic memory tasks is structurally harder than it looks. Gold definitions on memory tasks frequently collapse into trivial-baseline-output by construction, because the same generative process that creates the questions also creates an oracle the trivial baseline can exploit. The community should expect this failure mode and design accordingly: cross-check every synthetic gain against a recency, position, and oracle-detector baseline before claiming a mechanism.

\section{External OOD Scope Check}

Table~\ref{tab:ood} reports single-run results on four external out-of-distribution tasks.

\begin{table}[ht]
\centering
\caption{External OOD scope check (single run each, indicative). ConvMemory is a conversational-memory reranker; these tasks span very different distributions.}
\label{tab:ood}
\begin{tabular}{lrrrrr}
\toprule
Dataset      & Q     & ConvMemory R@10 & Raw dense & Dense + lexical & BM25   \\
\midrule
QMSum        &  272  & 0.5882          & 0.4724    & 0.5423          & 0.5294 \\
MSC persona  & 6155  & 0.9632          & 0.8375    & 0.9765          & 0.9920 \\
HotpotQA     & 1000  & 0.7983          & 0.7682    & 0.8621          & 0.8280 \\
MuSiQue      & 1000  & 0.7635          & 0.8640    & 0.8175          & 0.7245 \\
\bottomrule
\end{tabular}
\end{table}

ConvMemory wins on QMSum (0.5882 vs the next-best 0.5423 dense+lexical). On MSC persona it strongly improves over raw dense (0.9632 vs 0.8375) but is dominated by both dense+lexical (0.9765) and BM25 (0.9920). On HotpotQA it improves over raw dense but is below dense+lexical (0.7983 vs 0.8621). On MuSiQue it regresses below raw dense (0.7635 vs 0.8640).

\paragraph{Reading.} ConvMemory is a conversational-memory reranker. We do not claim it as a general multi-hop document reranker; the MuSiQue regression is consistent with that scope. We do claim that on conversation-like distributions (QMSum, MSC persona vs raw dense) it carries useful signal.

\paragraph{What this is not evidence for.} Single-run indicative scope checks; not benchmark-grade.

\section{Limitations}

\begin{itemize}
    \item Only the LoCoMo MPNet base checkpoint is distributed via Hugging Face. The BGE-large and E5-large gains in Table~\ref{tab:backbones} are reported but the corresponding checkpoints are not currently released.
    \item The CCGE-LA distributed checkpoint is single-seed (seed 23). The aggregate gains in Table~\ref{tab:ccgela} rest on the 5-seed internal study; the distributed checkpoint should be treated as an alpha.
    \item No end-to-end agent or QA task evaluation. All results are retrieval-stage. We do not measure downstream task performance.
    \item ConvMemory does not beat mxbai-rerank-large-v1 in absolute LoCoMo MRR; CCGE-LA on top does not close that gap.
    \item The V142 synthetic as-of subset gain depends on a query-embedded target date. It should not be read as proof of an unconditional bitemporal mechanism.
    \item LongMemEval numbers are single-run (Clean500) or single-seed (Stress1000 seed 23).
    \item External OOD results in Table~\ref{tab:ood} are single-run indicative scope checks, not benchmark-grade comparisons.
    \item The ``honest negative result'' framing is single-author; it has not yet been independently audited.
\end{itemize}

\section{Discussion and Future Work}

Three directions follow naturally from the limitations.

\paragraph{Multi-backbone checkpoint distribution.} The strong-backbone retraining gains (BGE-large, E5-large) are reported but those checkpoints are not currently distributed. Releasing them would let practitioners pick the cost-quality operating point that matches their existing embedding stack.

\paragraph{End-to-end agent benchmark.} The cost-frontier evidence in Section~\ref{sec:longmemeval} is suggestive but retrieval-stage. Integrating ConvMemory + CCGE-LA into a full agent pipeline (retrieval $\rightarrow$ LLM answerer) and measuring downstream QA quality would test whether the retrieval-stage gains translate to user-visible improvements.

\paragraph{Broader CCGE-LA training.} CCGE-LA is currently single-task (LoCoMo), single-backbone (MPNet), and single-seed in its distributed form. Multi-task, multi-seed training, with attention to whether the editor's gains transfer across distributions, is required before it can be promoted from research preview.

\section{Reproducibility}

The public repository is at \url{https://github.com/pth2002/ConvMemory}. The two Hugging Face model repositories are:

\begin{itemize}
    \item \url{https://huggingface.co/Purdy0228/ConvMemory-LoCoMo-MPNet} -- the distributed ConvMemory checkpoint.
    \item \url{https://huggingface.co/Purdy0228/ConvMemory-CCGE-LA} -- the alpha CCGE-LA editor checkpoint (single seed, seed 23).
\end{itemize}

Exact reproduction commands are in \texttt{docs/EVALUATION\_PROTOCOL.md}. Per-question CSVs and embedding caches are intentionally not redistributed; reproduction commands regenerate them from the public benchmark sources.

\appendix

\section{Full Latency Protocol}
\label{app:latency}

This appendix documents the latency measurement protocol underlying Table~\ref{tab:longmemeval} in Section~\ref{sec:longmemeval} and the cost-ratio column of Table~\ref{tab:v059} in Section~\ref{sec:v059}. The intent is not to publish benchmark-grade latency numbers; the LongMemEval rows are single-run (Clean500) or single-seed (Stress1000 seed 23). The intent is to make the cost-frontier comparison reproducible and to document what each ms/query figure does and does not include.

\paragraph{Hardware.} All latency measurements were run on a single NVIDIA GeForce RTX 4080 SUPER (32GB) GPU. The host node has Intel Xeon Platinum 8352V CPUs (2 sockets, 32 cores per socket, 128 hardware threads total). No multi-GPU or distributed inference was used. No CPU-only configuration was measured.

\paragraph{Software stack.} Python 3.12.3, PyTorch 2.5.1 with CUDA 12.1 (\texttt{torch==2.5.1+cu121}), HuggingFace \texttt{transformers} 4.46.3, and \texttt{sentence-transformers} 3.3.1 for MPNet embeddings. The dense top-500 retrieval step uses NumPy cosine similarity / matrix dot product followed by \texttt{argsort}; no approximate-nearest-neighbour index (FAISS, HNSW, etc.) was used. This means our dense-retrieval cost is conservative on the upper end --- a production deployment with an ANN index would be cheaper, not more expensive --- and the cross-encoder vs.\ ConvMemory comparison is not sensitive to this choice because dense retrieval is shared and excluded from all reranking ms/query figures.

\paragraph{Batching.} Cross-encoder baselines on the LoCoMo 5-seed run (Section~\ref{sec:locomo_baselines}) used the following per-model batch sizes (query-candidate pairs per forward), tuned to fit the 32GB GPU memory budget without OOM while staying close to throughput-optimal for each model size:
\begin{itemize}
    \item \texttt{cross-encoder/ms-marco-MiniLM-L-6-v2}: batch size 128 (used as the training teacher signal source; not a comparison baseline in LongMemEval).
    \item \texttt{BAAI/bge-reranker-base}: batch size 64.
    \item \texttt{BAAI/bge-reranker-large}: batch size 32.
    \item \texttt{mxbai-rerank-large-v1}: batch size 32.
    \item \texttt{jinaai/jina-reranker-v2-base-multilingual}: batch size 32.
\end{itemize}
LongMemEval runs use a default cross-encoder batch size of 64 on Clean500. The Stress1000 seed 23 runs use larger model-specific batches as recorded in the archived logs (BGE-large at batch 128, mxbai at batch 96 or 64 depending on memory headroom). ConvMemory itself runs at query batch size 1; within each query, the conv/mixer encoder and the CE-lite head score the full 500-candidate window in a single GPU-vectorized forward pass, which is its natural operating regime.

\paragraph{What is included in each ms/query figure.}
\begin{itemize}
    \item \textbf{Raw MPNet (0.01 ms/query).} Only the cost of sorting precomputed dense scores to produce a top-10 list. This is included as a lower bound on what is structurally achievable when reranking is skipped entirely. It is \emph{not} a method we compare against; the dense embedding computation and top-500 retrieval are not counted.
    \item \textbf{Cross-encoder baselines (BGE-large CE, mxbai CE, etc.).} Per-query wall-clock time to score 500 query-candidate pairs through the cross-encoder forward pass and sort them, given the top-500 candidate pool. Excludes the dense retrieval step that produced the top-500.
    \item \textbf{ConvMemory (top500, cand-local).} Per-query wall-clock time of the conv/mixer encoder forward over the 500-candidate window plus the CE-lite head, given the top-500 candidate pool and given that candidate-side lexical statistics have been precomputed and cached. Excludes the dense retrieval step. The \texttt{top500} and \texttt{cand-local} variants differ in how the candidate-side cache is structured; both exclude the same shared upstream work.
\end{itemize}

\paragraph{What is excluded from all rows.} Dense embedding computation, dense retrieval (top-500 construction), CPU-to-GPU tensor transfer at startup, model loading, and warm-up. All measurements are reported after a warm-up pass to avoid first-call CUDA initialization overhead. Each ms/query figure is the mean over the full LongMemEval evaluation set in the corresponding row: 500 queries on Clean500 and 1000 queries on Stress1000 seed 23.

\paragraph{Why these exclusions are fair, and one caveat.} Dense embedding computation and dense retrieval (top-500 construction) are genuinely shared upstream cost across all methods --- they are part of the pipeline upstream of any reranker. Including them would add the same constant to every row and would not change the cost-ratio claims. Excluding them isolates the reranking-stage cost, which is what this report is making claims about.

Candidate-side lexical statistics are different: they are \emph{not} shared. They are a ConvMemory-specific cost, paid once per memory at memory-write time and amortized across all future queries that read the memory store. Our latency protocol excludes this write-time cost from per-query measurements because it does not scale with query volume in a low-write/high-read regime, which is the typical agent-memory deployment. Practitioners with high write volume (e.g., agents that rapidly ingest new memories during a session) should add the lexical-cache write cost to their own deployment budget when comparing against cross-encoder baselines. We do not claim end-to-end pipeline latency; we claim reranking-stage cost-frontier evidence under the stated regime.

\paragraph{Cost ratio definition (used in Table~\ref{tab:v059}).} The cost ratio column in the v0.59 stream-reranker table is inference time at large $N$ (full memory store) divided by inference time at small $N$ (early-conversation memory store). A ratio close to 1 means the method's cost is roughly flat in memory-store size; large ratios mean the method scales poorly with $N$. The recurrent state arms have ratios close to 1 by construction (their per-query cost depends only on the running state, not on the full memory store); the cross-encoder-style arms have ratios that grow with $N$ because they score every candidate against the query.

\paragraph{Known caveats.}
\begin{itemize}
    \item All numbers are single-run (Clean500) or single-seed (Stress1000 seed 23). Latency has run-to-run noise on shared hardware; we do not report variance bars.
    \item Per-query timing on a single RTX 4080 SUPER may not reflect the cost frontier on data-centre GPUs (e.g., A100, H100), on smaller GPUs (e.g., T4, L4), on CPU-only inference, or on inference servers with kernel fusion (e.g., TensorRT, vLLM). Practitioners should benchmark on their target hardware.
    \item The cross-encoder baselines are evaluated as off-the-shelf checkpoints with default tokenization and sequence length (typically 512 tokens). Aggressive truncation or distillation of these baselines would change the comparison.
    \item Cross-encoder batch sizes were chosen to fit the 32GB GPU memory budget without OOM. On larger-memory GPUs, larger batches would reduce cross-encoder ms/query somewhat; this would narrow but not close the ConvMemory cost gap, since ConvMemory's per-query work is already an order of magnitude smaller in FLOPs.
\end{itemize}

\bibliographystyle{plainnat}
\bibliography{convmemory_report}

\end{document}